\documentclass[lettersize,journal]{IEEEtran}
\usepackage{amsmath,amsfonts}
\usepackage{algorithmic}
\usepackage{algorithm}
\usepackage{array}
\usepackage[caption=false,font=normalsize,labelfont=sf,textfont=sf]{subfig}
\usepackage{textcomp}
\usepackage{stfloats}
\usepackage{url}
\usepackage{verbatim}
\usepackage{graphicx}
\usepackage{cite}
\usepackage{hyperref}
\usepackage{enumitem}
\usepackage{chngcntr}
\begin{document}

\title{%
  {\fontsize{18}{22}\selectfont
   LiteVoxel: Low-memory Intelligent Thresholding
for Efficient Voxel Rasterization }%
}

\author{%
  Jee Won Lee\textsuperscript{1,2}, 
  and Jongseong Brad Choi\textsuperscript{1,2*}\\[0.3ex]
  {\footnotesize\normalfont%
    \textsuperscript{1} Department of Mechanical Engineering, State University of New York, Korea, Incheon, South Korea\\[0.3ex]
    \textsuperscript{2} Department of Mechanical Engineering, State University of New York, Stony Brook, NY, United States\\[0.3ex]
    \textsuperscript{*}Corresponding author
  }%

\begin{center}
  \includegraphics[width=0.9\textwidth]{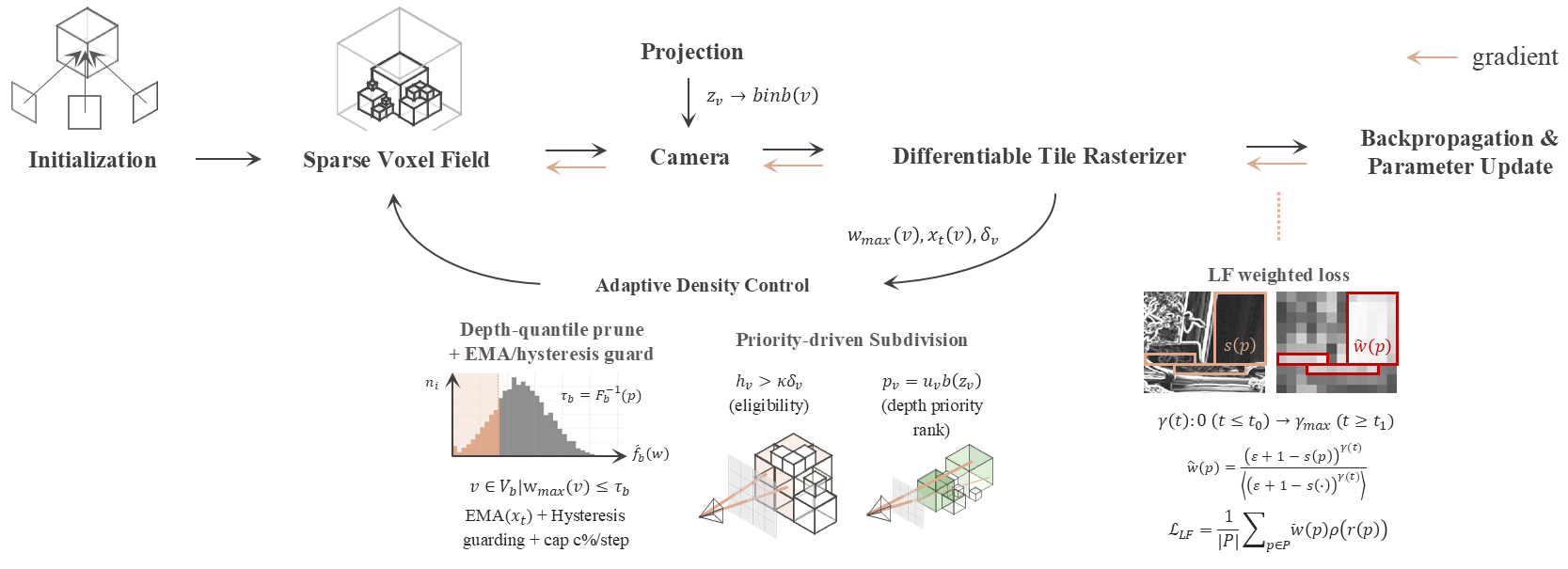}\\[0.5ex]
  {\footnotesize
   \begin{minipage}{\textwidth}
     \raggedright
     \textbf{Fig.~1. LiteVoxel Overview.} Our system trains a sparse voxel field with a differentiable tile rasterizer and three targeted innovations. (i) LF-weighted loss: a Sobel-inverse weight with a mid-training $\gamma$-ramp shifts gradient to low-texture regions without harming edges. (ii) Depth-quantile pruning: per-depth bins prune voxels with low max blending weight $w_{\mathrm{max}}$ using a quantile cutoff $\tau_b$, stabilized by EMA + hysteresis and a small per-step cap. (iii) Priority-driven subdivision: ray-footprint eligibility $(h_v > \kappa \, \delta_v)$ and a depth-aware priority $p_v = u_v \, b(z_v)$ select splits under a growth budget with optimizer re-initialization. Together, these modules bound VRAM and model size while improving low-frequency fidelity and boundary stability.
   \end{minipage}
  }
\end{center}
\thanks{This work was supported by the National Research Foundation of Korea (NRF) grant funded by the Korea government (MSIT) (Grant No. RS-2022-NR067080 and RS-2025-05515607) (Corresponding author: Jongseong Brad Choi). 
Jee Won Lee: Data Curation, Methodology, Validation, Visualization, Writing – original draft, Software. 
Jongseong Brad Choi: Conceptualization, Methodology, Validation, Formal Analysis, Writing - Review \& Editing.}%
\thanks{Jee Won Lee is with the State University of New York City Korea, Incheon 21985, South Korea (jeewon.lee@stonybrook.edu). 
Jongseong Brad Choi is with the Department of Mechanical Engineering, State University of New York City Korea, Incheon 21985, Korea (jongseong.choi@stonybrook.edu).}%
}

\maketitle

\begin{abstract}
Sparse-voxel rasterization is a fast, differentiable alternative for optimization-based scene reconstruction, but it tends to underfit low-frequency content, depends on brittle pruning heuristics, and can overgrow in ways that inflate VRAM. We introduce LiteVoxel, a self-tuning training pipeline that makes SV rasterization both steadier and lighter. Our loss is made low-frequency aware via an inverse-Sobel reweighting with a mid-training $\gamma$-ramp, shifting gradient budget to flat regions only after geometry stabilize. Adaptation replaces fixed thresholds with a depth-quantile pruning logic on maximum blending weight, stabilized by EMA-hysteresis guards and refines structure through ray-footprint-based, priority-driven subdivision under an explicit growth budget. Ablations and full-system results across Mip-NeRF~360 (6~scenes) and Tanks~\&~Temples (3~scenes) datasets show mitigation of errors in low-frequency regions and boundary instability while keeping PSNR/SSIM, training time, and FPS comparable to a strong SVRaster baseline. Crucially, \textbf{LiteVoxel} reduces peak VRAM by $\sim$40--60\% and preserves low-frequency detail that prior setups miss, enabling more predictable, memory-efficient training without sacrificing perceptual quality.

\textit{Index Terms}—Image reconstruction, Image synthesis, Octrees, Structure from Motion, Thresholding (Imaging)

\end{abstract}

\section{Introduction}
\IEEEPARstart{V}{IEW}-synthesis and reconstruction surged with radiance-field methods such as NeRF by Mildenhall et al.~\cite{ref1}, but training/rendering speed and memory pressure motivated a shift toward faster, explicit alternatives. 3D Gaussian Splatting (3DGS), introduced by Kerbl et al.~\cite{ref2}, achieves real-time rendering by optimizing anisotropic Gaussian primitives with a visibility-aware rasterizer, catalyzing a wave of explicit, differentiable pipelines that trade deep MLPs for direct parameter optimization. This trend invites a re-examination of classical voxel structures as practical, differentiable substrates for efficient pipelines.

\indent Voxel and sparse voxel octrees, in particular, have long served graphics and reconstruction for their visibility handling and hierarchical culling~\cite{ref3}. The depth-sensor era popularized dense Truncated Signed Distance Function (TSDF) grids and voxel hashing for large-scale mapping~\cite{ref4},~\cite{ref5}. However, dense memory footprints, aliasing, and level-of-detail control limited adoption in high-fidelity view synthesis. Sparse Voxels Rasterization (SVRaster) proposed by Sun et al., revisits voxels with a differentiable rasterizer and adaptive sparsity, achieving high quality and speed without neural networks and positioning voxels as a viable alternative to Gaussians and neural fields~\cite{ref6},~\cite{ref7}.

\indent While NeRF and explicit rasterization pipelines such as 3DGS have matured rapidly, the community has continued to address reconstruction blind spots by their novel techniques. Micro-Splatting by Lee et al.~\cite{ref8} focuses on compact, near-isotropic primitives and disciplined densification, AbsGS by Ye et al.~\cite{ref9} analyzes gradient interactions and proposes homodirectional view-space gradients to stabilize splitting, and Gaussian Surfels by Dai et al.~\cite{ref10} align primitives to local surfaces for sharper geometry. As complementary efforts refine Gaussian pipelines, our focus is the voxel regime. We target the failure modes specific to sparse-voxel rasterization and design mechanisms native to voxels to improve low-frequency fidelity, depth fairness, and sparsity stability.

\indent Despite recent progress, baseline SVRaster suffers (i) low-frequency underfitting (blotchy flats), (ii) depth-biased pruning with silhouette flicker, and (iii) myopic subdivision that over-refines near geometry and ignores far voxels, inflating VRAM. Therefore, we introduce LiteVoxel, a self-tuning training pipeline for sparse-voxel rasterization that achieves steadier optimization and lower memory, directly addressing the practical gaps of the baseline SVRaster pipeline. Our contributions are as follows: 

\indent 

\begin{itemize}
 \item We introduce a low-frequency-aware photometric loss that reweights pixels via an inverse-Sobel map with a mid-training $\gamma\!\left(t\right)$ ramp and mean normalization, redirection gradient to flat regions only after geometry stabilizes thus removing blotchy planes without sacrificing details or PSNR/SSIM/runtime. 
 \item We propose depth-quantile pruning with stability guards, per-depth cumulative distribution function cutoffs on maximum blending weight combined with EMA-hysteresis, contour-dilation, and a per-step cap, which adapts pruning fairly across depth to eliminate boundary flicker while reducing model size and peak VRAM. 
 \item We develop priority-driven subdivision that uses a ray-footprint eligibility test that ranks eligible voxels by usefulness with a mild far-depth bias which selects splits under a global budget with optimizer re-initialization, ensuring refinement where cameras can resolve detail, attention to far regions, and bounded growth (\textasciitilde40\%-60\% lower peak VRAM) with comparable FPS and training time.

\end{itemize}

\section{Related Work}
\subsection{Sparse-Voxel Rasterization (SVRaster)}
\indent SVRaster reconstructs and renders a scene by directly optimizing a sparse voxel field with a differentiable tile rasterizer and a photometric objective without neural networks. Rays are traced from camera centers through pixel footprints where the rasterizer accumulates per-voxel color/opacity along each ray, and gradients from the image loss flow directly to voxel parameters~\cite{ref6}. The adaptation loop periodically prunes unhelpful voxels and subdivides useful ones to match image resolution, yielding real-time rendering with competitive fidelity.

\indent We adopt SVRaster’s native statistics that drive adaptation: 
the maximum blending weight $w_{\mathrm{max}}(v)$ (a proxy for voxel contribution/visibility), and the local inter-ray sample interval $\delta_v$ (the camera-induced footprint at the voxel’s depth). These signals are inexpensive to gather during rasterization and are central to our redesign of pruning and subdivision.

\subsection{Fidelity Enhancements in Explicit Rasterization}
\indent 3D Gaussian Splatting (3DGS), introduced by Kerbl et al., demonstrated that real-time view synthesis is attainable with explicit, rasterizable primitives optimized directly in image space, catalyzing a family of non-neural, differentiable pipelines~\cite{ref2}. Building on 3DGS, several methods explicitly target missed detail via improved densification and geometry alignment. Micro-Splatting by Lee et al. emphasizes compact, near-isotropic primitives and disciplined densification to recover fine structure~\cite{ref8}. AbsGS by Ye et al. analyzes gradient interactions in the splatting objective and proposes homodirectional, view-space gradients to stabilize splitting and improve high-frequency recovery~\cite{ref9}. Gaussian Surfels by Dai et al. align anisotropic primitives to local surfaces (surfels) for sharper geometry and crisper edges~\cite{ref10}. 
\\\indent Recent work on explicit rasterization shows that careful loss weighting and adaptive refinement can remedy reconstruction blind spots. Motivated by these insights, we target sparse-voxel rasterization and introduce a voxel-native training pipeline that targets the baseline’s overlooked issues in low-frequency fidelity, pruning stability, and refinement policy. Inspired by these advances, we target sparse-voxel rasterization directly, introducing mechanisms native to voxels rather than porting Gaussian-specific tricks.

\subsection{Low-Frequency Underfitting in SVRaster}
\indent Low-frequency regions are systematically underfit in sparse-voxel rasterization. Although SVRaster preserves high-frequency detail, the training signal is strongly edge-driven as pixelwise photometric losses produce large gradients at contours and high-contrast areas, while low-texture regions receive weak, noisy supervision, leading to blotchy planes~\cite{ref6},~\cite{ref7}. This imbalance is compounded along the ray because flat-region errors are spatially diluted, spread over many pixels, so each voxel explains only a tiny share of the bias. Also, since accumulated transmittance in large, smooth surfaces often saturate near zero or one, both effects make the derivative of the rendered color with respect to opacity nearly vanish and further weakens gradients in flats. 

\indent Classical remedies help but don’t directly fix this LF underfitting. SSIM (Structural Similarity Index Measure) is routinely paired with L1/L2 to better match perception, and robust penalties stabilize optimization under exposure drift and outliers~\cite{ref11},~\cite{ref12}. Appearance/exposure handling in view-synthesis reduces bias from brightness shifts~\cite{ref1},~\cite{ref13}. However, these components still emphasize structure or merely damp outliers as they do not explicitly reallocate gradient budget toward flat regions in a voxel rasterizer. 
\\\indent Motivated by this gap, we introduce a low-frequency-aware reweighting that increases the contribution of flat pixels only after early geometry has stabilized, counteracting the edge bias while preserving high-frequency sharpness. 

\subsection{SVRaster VRAM Inflation}
\indent Despite its efficiency at render time, a vanilla sparse-voxel rasterizer often inflates memory during training. Each occupied voxel carries a nontrivial parameter set (position/extent and several SH color coefficients) plus optimizer state (e.g., Adam’s moments), so footprint scales linearly with voxel count~\cite{ref6},~\cite{ref14}. Early iterations favor aggressive octree growth: uniform or edge-driven splitting multiplies cells by up to 8× per step and is biased toward near-camera regions where rays are densest. Since pruning is typically governed by a global threshold on per-voxel blending weight, it is depth-biased being conservative on far voxels and over-zealous near silhouettes so many low-value cells linger while useful far cells are retained unevenly~\cite{ref6},~\cite{ref7}. Around object boundaries, noisy inside/outside estimates cause “keep halos,” further bloating the grid. Peak usage also reflects transient buffers needed for tiling/compositing and statistic passes, which grow with the number of live voxels. These behaviors together create large, spiky VRAM peaks even when final reconstructions could be represented far more compactly thereby motivating our depth-aware pruning and budgeted, footprint-tested subdivision to keep growth in check without sacrificing quality.

\section{Preliminaries}
\noindent\textit{Camera and Rays.} 
\\\indent Throughout, voxels are indexed by $v$ and pixels by image coordinates $\left(x, y\right)$. 
Let $\mathbf{o} \in \mathbb{R}^{3}$ be the camera center and $\widehat{\mathbf{d}}(x, y) \in \mathbb{S}^{2}$ 
the unit direction through pixel $(x, y)$ as shown in Fig.~2. 
A primary ray is 
\begin{equation}
\mathbf{r}(t) = {o} + t\,{d}(x, y), \quad t \ge 0.
\end{equation}

\noindent\textit{Sparse Voxel/Octree Indexing.}
\\\indent The scene is discretized into cube voxels $v$ with center $\mathbf{x}_{v}$, 
half-size $h_{v}$, RGB color $\mathbf{c}_{v} \in [0, 1]^{3}$, 
and opacity $a_{v} \in [0, 1]$. Voxels are organized in an octree; 
level $l$ voxel cells have edge length $2^{l} h_{0}$, and $l_{\mathrm{max}}$ denotes the finest level.

\vspace{1\baselineskip}

\noindent\textit{Differentiable compositing (per ray).}
\\\indent Let $\left\{ v_k \right\}_{k=1}^{K}$ be the voxels intersected by $\mathbf{r}(t)$, ordered front-to-back. Using standard alpha compositing,

\begin{figure}[H]        
  \centering
  \includegraphics[width=0.85\linewidth]{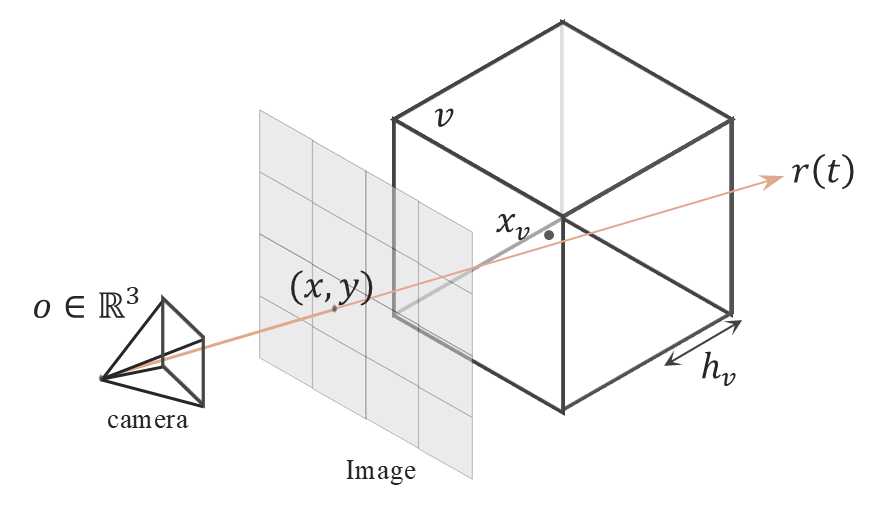}\\[2ex]
  {\footnotesize
   \parbox{\columnwidth}{
     \textbf{Fig.~2. \textit{Voxel Parameters and Ray Geometry.}} 
     A pinhole camera at $\mathbf{o}$ casts a ray 
     $\mathbf{r}(t) = \mathbf{o} + t\,\mathbf{d}(x, y)$ 
     through pixel $(x, y)$ toward a voxel $v$ 
     with center $\mathbf{x}_{v}$ and half-size $h_{v}$.
   }
  }
\end{figure}

\begin{gather}
T_{0} = 1, \quad w_{k} = T_{k-1}\,\alpha_{v_k}, \tag{2}\label{eq:alpha1}\\[4pt]
T_{k} = T_{k-1}\!\left(1 - \alpha_{v_k}\right), \tag{3}\label{eq:alpha2}\\[4pt]
\widehat{C}(x, y) = \sum_{k=1}^{K} w_{k}\,\mathbf{c}_{v_k}. \tag{4}\label{eq:alpha3}
\end{gather}

Here, $w_{k}$ is the contribution of voxel $v_{k}$, 
$\alpha_{v_k}$ its opacity, and $T_{k}$ the accumulated transmittance up to the $k$-th sample.

\vspace{1\baselineskip}

\noindent\textit{Per-voxel statistics used for adaptation.}
\\\indent We employ the maximum blending weight observed for voxel $v$,
\begin{equation}
w_{\mathrm{max}}(v) = \max_{\text{rays } r \text{ hitting } v} 
\left\{ T_{k(v, r)-1}\,\alpha_{v} \right\}, 
\tag{5}\label{eq:wmax}
\end{equation}
And the local inter-ray spacing $\delta_{v}$ 
(the image-plane footprint at $\mathbf{x}_{v}$) as a depth proxy.

\vspace{1\baselineskip}

\noindent\textit{Loss Optimization.}
\\\indent Given the rendered color $\widehat{C}(x, y)$ and ground truth $I(x, y)$, 
the standard sparse-voxel rasterization loss is defined as
\begin{equation}
\begin{split}
L_{\mathrm{SVR}} =\;&
L_{\mathrm{mse}}
+ \lambda_{\mathrm{ssim}} L_{\mathrm{ssim}}
+ \lambda_{T} L_{T} \\
&+ \lambda_{\mathrm{dist}} L_{\mathrm{dist}}
+ \lambda_{R} L_{R}
+ \lambda_{\mathrm{tv}} L_{\mathrm{tv}}.
\end{split}
\tag{6}\label{eq:svr-loss}
\end{equation}

Here, $L_{\mathrm{mse}}$ is the pixelwise MSE; 
$L_{\mathrm{ssim}}$ measures structure similarity; 
$L_{T}$ promotes near-binary transmittance; 
$L_{\mathrm{dist}}$ is a distortion term in ray space; 
$L_{R}$ regularizes per-voxel color; 
and $L_{\mathrm{tv}}$ is the total variation on the density/opacity grid. 
Section~IV-E specifies our modifications to the photometric term 
and the subset of auxiliaries used in LiteVoxel.

\section{Methodology}
\subsection{Overview}
\indent LiteVoxel is a principled training framework for sparse-voxel rasterization that rethinks both supervision and capacity control within the classic analysis-by-synthesis loop (Fig. 1). Starting from an SfM-initialized voxel field, we project, rasterize, measure a photometric objective, and back-propagate. Our contributions center on two coordinated components that address fundamental failure modes of voxel rasterization—edge-dominated gradients, depth-coupled sampling bias, and unstable sparsity updates—without adding neural modules or heavy priors. Together, these components yield a self-tuning pipeline that preserves the simplicity and speed of voxel rasterization while substantially reducing peak VRAM and improving reconstruction of low-frequency structure. 

\subsection{Low-frequency-aware photometric reweighting}
\indent Edges naturally dominate the photometric gradient in voxel rasterization, leaving large, flat regions under-optimized. We counter this by redistributing per-pixel gradients with an inverse-Sobel weight~\cite{ref15}, following the edge-aware weighting principle used to attenuate gradients near high-contrast boundaries~\cite{ref16},~\cite{ref18}. The reweighting is activated mid-training as a simple curriculum to avoid disturbing early geometric convergence ~\cite{ref17} and is mean-normalized to prevent loss-scale drift (Fig. 3).
\\\indent Let $s(p) \in [0,1]$ be the percentile-normalized Sobel magnitude at pixel $p$, treated as stop-grad. 
We form the unnormalized weight
\begin{equation}
w(p) = (\varepsilon + 1 - s(p))^{\gamma(t)}, \quad \varepsilon \approx 10^{-3}.
\tag{7}\label{eq:lfw}
\end{equation}

\noindent and stabilize the overall gradient scale by mean normalization.
\begin{equation}
\widetilde{w}(p) = 
\frac{w(p)}{\langle w(\cdot) \rangle}, \quad
\langle w(\cdot) \rangle = 
\frac{1}{|P|} \sum_{q \in P} w(q).
\tag{8}\label{eq:lfwnorm}
\end{equation}

\noindent The LF emphasis factor $\gamma\!\left(t\right)$ follows a piecewise-linear schedule
\begin{equation}
\gamma(t) =
\begin{cases}
0, & t < t_{0},\\[4pt]
\gamma_{\max} \frac{t - t_{0}}{t_{1} - t_{0}}, & t_{0} \le t < t_{1},\\[4pt]
\gamma_{\max}, & t \ge t_{1},
\end{cases}
\tag{9}\label{eq:gamma-schedule}
\end{equation}

with default $t_{0} \approx 0.3T$, $t_{1} \approx 0.6T$, $\gamma_{\max} \approx 0.6$. 
Given residual $r_{p} = R(p) - G(p)$, the photo term becomes
\begin{equation}
L_{\mathrm{LF}} = 
\frac{1}{|P|} \sum_{p \in P} 
\widehat{w}(p) \, \rho(r_{p}).
\tag{10}\label{eq:lf-loss}
\end{equation}

where $\gamma(t) = 0$, $L_{\mathrm{LF}}$ reduces exactly to the baseline photometric loss. 
As $\gamma(t)$ ramps up mid-training, gradients on high-Sobel (edge) pixels are relatively down-weighted 
while those on low-Sobel (flat) pixels are amplified. 
Intuitively, edges and textures converge early where flats lag and exhibit blotchy residuals. 
The delayed, mean-normalized inverse-Sobel weighting increases attention to flat regions only after geometry stabilize, suppressing low-frequency blotching without softening edges, and keeps learning dynamics (step sizes, wall-clock) comparable to the baseline.

\begin{figure}[b]
  \centering
  \includegraphics[width=0.85\linewidth]{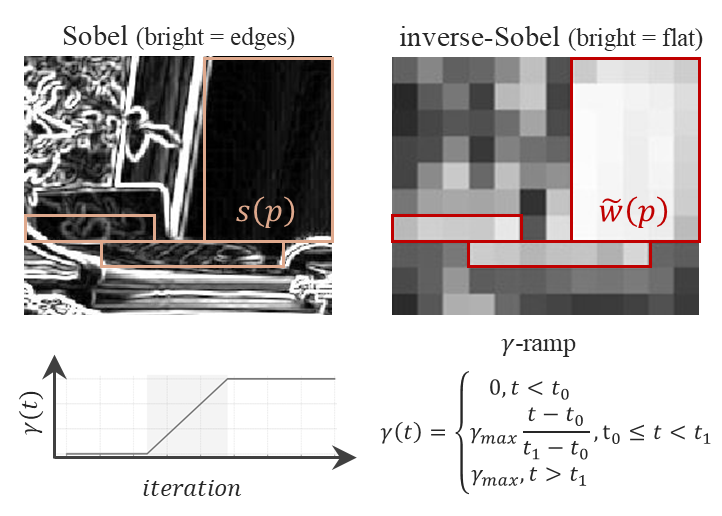}\\[2ex]
  {\footnotesize
   \parbox{\columnwidth}{
     \textbf{Fig.~3. \textit{Low-frequency-aware photometric reweighting.}} 
     From the inverse-Sobel weight map, the loss up-weights low-frequency regions; 
     a mid-training $\gamma\!\left(t\right)$-ramp gradually enables this emphasis, 
     with weights mean-normalized and the Sobel map treated as stop-grad to avoid.
   }
  }
\end{figure}
\subsection{Depth-aware quantile pruning}
\indent To mitigate the baseline’s peak-VRAM spikes caused by uniform growth near cameras and a single global prune threshold that under-removes far voxels, we replace ad-hoc deletion with a depth-aware, adaptive rule. Voxels are pruned by per-depth quantiles of their maximum blending weight, and the decision is stabilized by an EMA–hysteresis inside test, a contour/detail “keep-halo,” and a per-step cap. The result is balanced sparsity across depth and stable silhouettes (see Fig. 4).

\indent We group voxels by a depth proxy into bins $V_{b}$. 
As a proxy we use the per-voxel maximum blending weights. 
The empirical CDF is
\begin{equation}
F_{b}(t) = 
\frac{1}{|V_{b}|}
\sum_{v \in V_{b}} 
\mathbf{1}\!\left\{ w_{v}^{\mathrm{max}} \le t \right\},
\tag{11}\label{eq:cdf}
\end{equation}
and the bin-wise threshold is the $q_{b}$-quantile,
\begin{equation}
\tau_{b} = F_{b}^{-1}\!\left(q_{b}\right).
\tag{12}\label{eq:quantile}
\end{equation}

Here, $q_{b} \in (0,1)$ is a per-bin quantile annealed over training with a mild near/far relaxation. 
We prune $v \in V_{b}$ when $w_{v}^{\mathrm{max}} \le \tau_{b}$. 
This allocates sparsity pressure fairly over depth and avoids the ``all-near, no-far'' bias of a single global cutoff. 
To avoid silhouette chatter and burst deletions, we add three guards:

\begin{enumerate}[leftmargin=1.8em, label={\roman*.}]
  \item \textit{EMA + hysteresis for inside labels.}
  Maintain an exponential moving average 
  $m_{t} = (1 - \alpha) m_{t-1} + \alpha x_{t}$ 
  of the inside score and update a discrete state 
  $s_{t} \in \{ \text{in}, \text{out} \}$ with hysteresis,
  \[
  s_{t} =
  \begin{cases}
  \text{out}, & m_{t} < m_{\mathrm{low}},\\[4pt]
  \text{in}, & m_{t} > m_{\mathrm{high}},\\[4pt]
  \text{otherwise}, & s_{t} = s_{t-1}
  \end{cases}
  \]
  This prevents flip-flops when $m_{t}$ hovers near the boundary.

  \item \textit{Contour/detail dilation (“keep-halo”).}
  Temporarily protect voxels that are small relative to the local ray footprint 
  or exhibit high $w_{v}^{\mathrm{max}}$ near edges where this preserves thin structures 
  and silhouette continuity.

  \item \textit{Per-step cap.}
  Limit the removable fraction at each adapt step, 
  turning potentially spiky deletion into smooth, predictable downsizing.
\end{enumerate}

\subsection{Priority-driven subdivision}
\indent To curb near-surface over-refinement, neglect of far geometry, and the resulting VRAM spikes in baseline sparse-voxel rasterization, we drive subdivision by an image-resolution test and a depth-aware priority.

\indent We permit splits only where image formation can justify extra capacity. 
For a voxel $v$ with half-size $h_{v}$, let $\delta_{v}$ denote the local inter-ray spacing, 
the projected ray “footprint”, at $v$’s location. 
A voxel is eligible if it is not at the finest octree level and
\begin{equation}
h_{v} > \kappa\,\delta_{v}, \quad \kappa \approx 1,
\tag{13}\label{eq:eligibility}
\end{equation}
which prevents refining regions that camera cannot resolve 
and removes a major source of runaway growth near the sensor (see Fig.~5).

Among eligible voxels, we assign a depth-aware priority that balances usefulness 
and a mild far-depth preference. 
Let $z_{v}$ be the camera-space depth of voxel $v$. 
We robust-normalize depth to $[0,1]$ using the running 5th/95th percentiles 
$\left(z_{\mathrm{p5}}, z_{\mathrm{p95}}\right)$:
\begin{equation}
\widetilde{z}_{v} =
\mathrm{clip}\!\left(
\frac{z_{v} - z_{\mathrm{p5}}}{z_{\mathrm{p95}} - z_{\mathrm{p5}}},
0, 1
\right).
\tag{14}\label{eq:depthnorm}
\end{equation}

\begin{figure}[t!]
  \centering
  \includegraphics[width=0.9\linewidth]{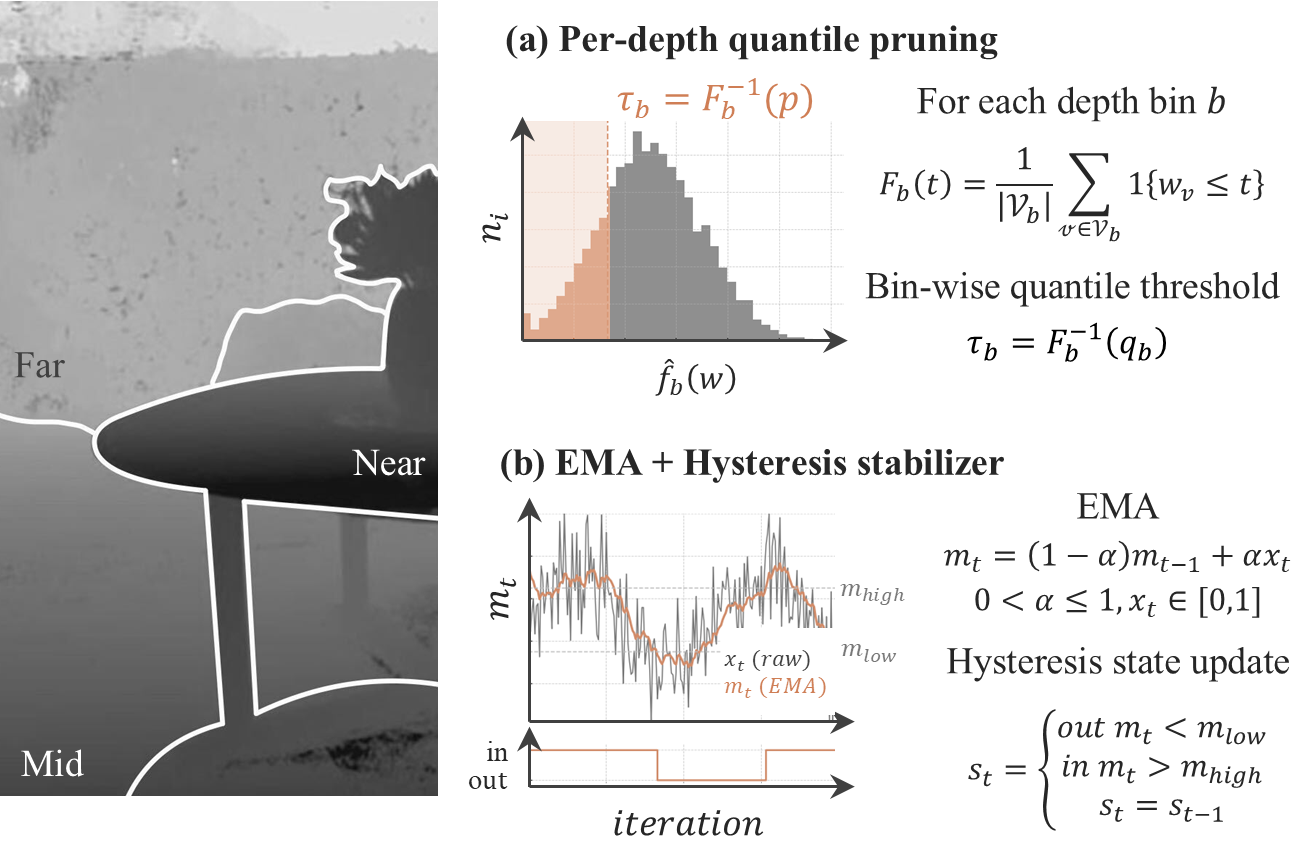}\\[2ex]
  {\footnotesize
   \parbox{\columnwidth}{
     \textbf{Fig.~4. \textit{Depth-aware pruning with stability guards.}} 
     (a) For each zone, voxels are pruned by a per-bin quantile of their maximum blending weight (orange region), 
     equalizing sparsity across depth. 
     (b) EMA + hysteresis converts noisy inside scores into stable in/out decisions, 
     preventing flip-flops along silhouettes.
   }
  }
\end{figure}

\noindent We then apply a gentle far-bias
\begin{equation}
b(z_{v}) = 1 + \beta\,\widetilde{z}_{v},
\tag{15}\label{eq:farbias}
\end{equation}
where $\widetilde{z}_{v} \in [0,1]$, 
$\beta \in [0.1,0.3]$, 
and scale each voxel’s running usefulness score $P_{v}$ by $b(\cdot)$:
\begin{equation}
P_{v} \leftarrow P_{v}\,b(z_{v}).
\tag{16}\label{eq:pvscale}
\end{equation}

\noindent This counters myopic splitting near the camera and ensures resolvable far regions receive attention 
as shown in Fig.~5, where darker green indicates higher $P_{v}$.

Finally, we perform budgeted selection where at each adaptation step, 
we choose only the top-$p_{v}$ candidates under a global growth cap (maximum splits per step) 
and refresh the optimizer/scheduler after topology changes. 
Together, footprint eligibility, depth-aware prioritization, and budgeted selection 
refine exactly where views support detail, allocate capacity across depth, 
and keep model growth and VRAM, predictable.

\subsection{LiteVoxel Training Objective}
Relative to the baseline SVRaster objective, we replace the photometric core with a low-frequency-weighted robust loss and retain only the auxiliary regularizer we actually use. The overall loss is

\begin{equation}
\begin{aligned}
L_{\mathrm{Total}} ={} &
L_{\mathrm{LF}}
+ \lambda_{\mathrm{SSIM}} L_{\mathrm{SSIM}}
+ \lambda_{T_{c}} L_{T_{c}}
+ \lambda_{T_{i}} L_{T_{i}} \\[3pt]
&+ \lambda_{\mathrm{NDm}} L_{\mathrm{NDm}}
+ \lambda_{\mathrm{NDM}} L_{\mathrm{NDM}} \\
&+ \lambda_{\mathrm{mask}} L_{\mathrm{mask}}
+ \lambda_{\mathrm{tv}} L_{\mathrm{tv}}.
\end{aligned}
\tag{17}\label{eq:total-loss}
\end{equation}

The LF-weighted photometric term is
\begin{equation}
L_{\mathrm{LF}}(t) =
\frac{1}{|P|}
\sum_{p \in P}
\widetilde{w}(p; t)\,
\rho\!\left(\widehat{C}(p) - I(p)\right),
\tag{18}\label{eq:lf-loss}
\end{equation}

where $P$ is the set of pixels, $\widehat{C}(p)$ is the rendered color, $I(p)$ is the ground truth, 
and $\rho(\cdot)$ is a robust penalty.

\begin{figure}[t!]
  \centering
  \includegraphics[width=0.9\linewidth]{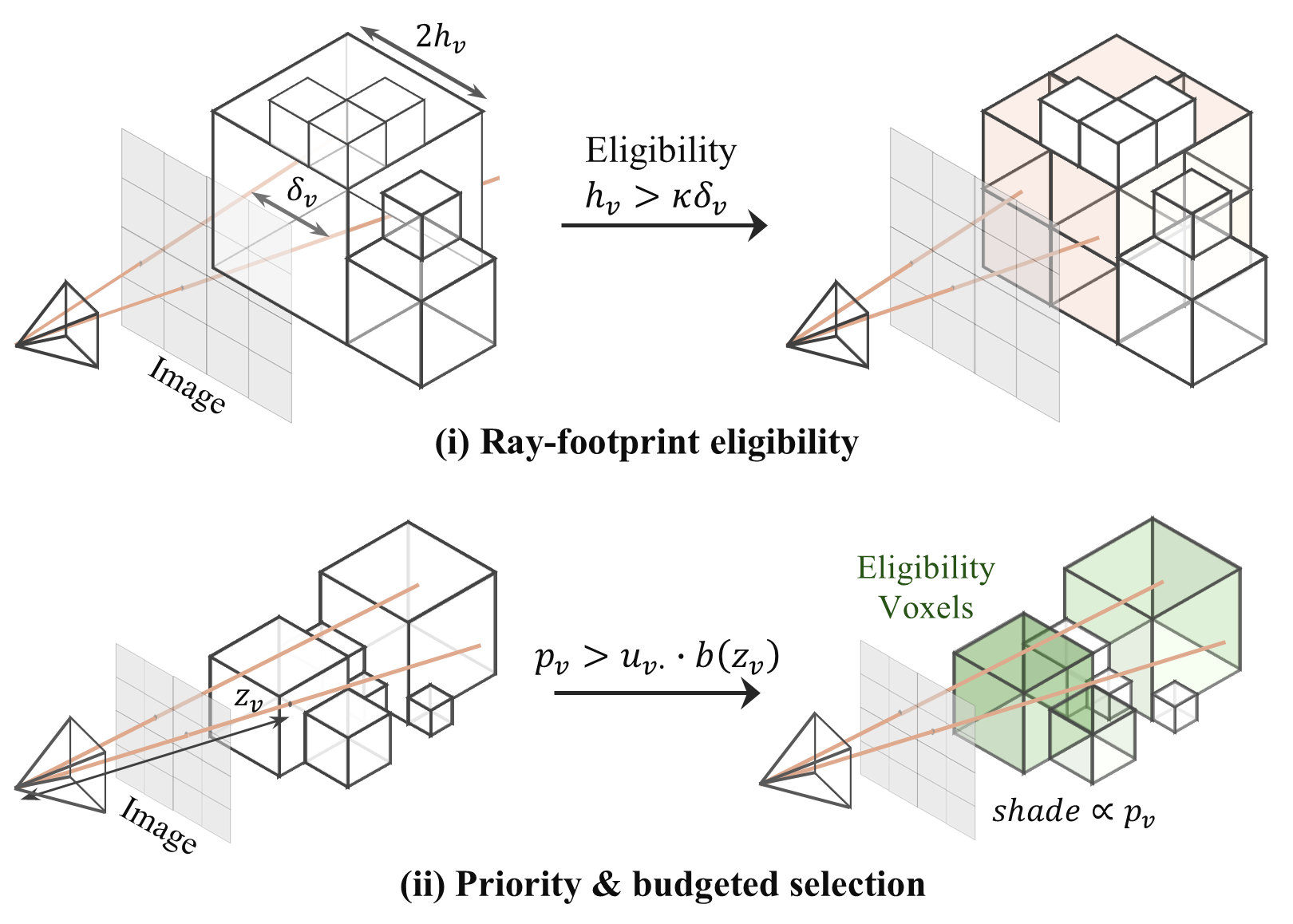}\\[2ex]
  {\footnotesize
   \parbox{\columnwidth}{
     \textbf{Fig.~5. \textit{Priority-driven subdivision with ray-footprint eligibility.}} 
     (i) Voxel becomes a candidate for splitting only if its half-size exceeds the local inter-ray spacing 
     on the image plane; this footprint test voids over-refining regions the camera cannot resolve. 
     (ii) Eligible voxels are ranked by a usefulness score with a mild depth bias; 
     the highest-priority candidates (green) are split under a global growth budget, 
     allocating refinement to resolvable near and far regions while keeping model size bounded.
   }
  }
\end{figure}

$L_{\mathrm{SSIM}}$ is the structure-aware penalty, 
$L_{T_{c}}$ is the transmittance concentration that encourages final ray transmittance to be near $0$ or $1$, 
$L_{T_{i}}$ is the inside regularizer on raw transmittance which discourages floating density and stabilizes empty space, 
$\lambda_{\mathrm{NDM}}$ is the depth-normal consistency to promote piecewise-smooth geometry, 
$L_{\mathrm{NDM}}$ is the median-based depth-normal consistency for robustness to outliers, 
$L_{\mathrm{mask}}$ is the foreground/background supervision, 
and $L_{\mathrm{tv}}$ is the total-variation on the sparse density field to reduce spurious high-frequency noise.

\indent LiteVoxel couples an LF-aware supervision scheme with depth-adaptive sparsity control—quantile pruning with stabilization and footprint-guided, budgeted subdivision—to keep capacity where images support detail while bounding growth. In the next section we evaluate these components quantitatively and qualitatively on standard view-synthesis benchmarks, reporting PSNR/SSIM, training/render throughput, peak VRAM, and ablations that isolate the effect of each module and the combined system.

\section{Results and Evalutaion}

We evaluate LiteVoxel on two standard view-synthesis benchmarks: Mip-NeRF 360~\cite{ref19} and Tanks \& Temples~\cite{ref20} (outdoor/large-scale scenes). These datasets stress both low-frequency expanses (skies, walls, floors) and high-frequency geometry (foliage, thin structures), which directly probe the three failure modes targeted by our method (LF underfitting, depth-biased pruning, and near-surface over-refinement). We report results on held-out views for each scene. 

\indent To isolate our contributions, we compare LiteVoxel against SVRaster~\cite{ref6} (the voxel rasterization baseline), and two explicit splatting methods, 3D Gaussian Splatting (3DGS)~\cite{ref2} and Micro-Splatting~\cite{ref8}, using public implementations and the same camera intrinsics and render settings where applicable. SVRaster and LiteVoxel share the same differentiable rasterizer and adaptation cadence. 

\begin{table*}[!t]
  \centering
  \caption{Quantitative Analysis Averaged Across Datasets}
  \includegraphics[width=\textwidth]{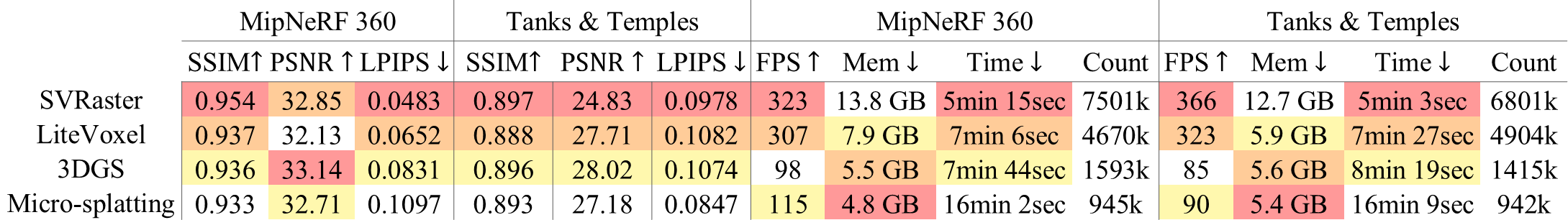}
  \label{tab:avg}
\end{table*}

\begin{figure*}[!t]
  \centering
  \includegraphics[width=\textwidth]{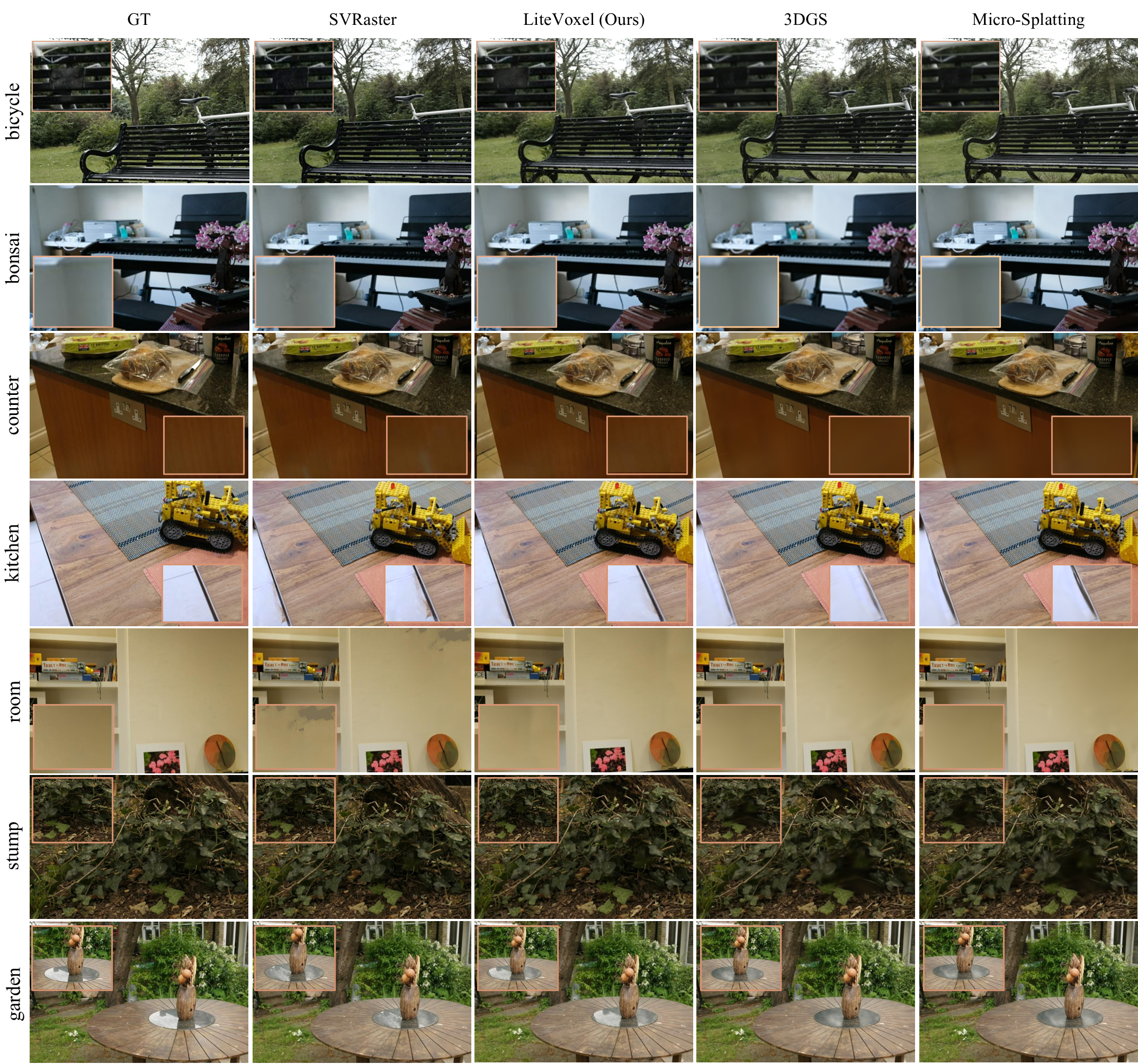}\\[2ex]
  {\footnotesize
  \parbox{\textwidth}{
    \textbf{Fig.~6. \textit{Qualitative comparison on Mip-NeRF~360~\cite{ref19}.}} 
    (From left to right) Each row shows the ground truth, SVRaster~\cite{ref6}, LiteVoxel (Ours), 3DGS~\cite{ref2}, and Micro-Splatting~\cite{ref8}. 
    Scenes such as \textit{bicycle}, \textit{bonsai}, \textit{counter}, \textit{kitchen}, 
    \textit{room}, \textit{stump}, and \textit{garden} demonstrate how our approach captures detail 
    and removes errors more effectively.
  }}
\end{figure*}

\begin{figure*}[!t]
  \centering
  \includegraphics[width=\textwidth]{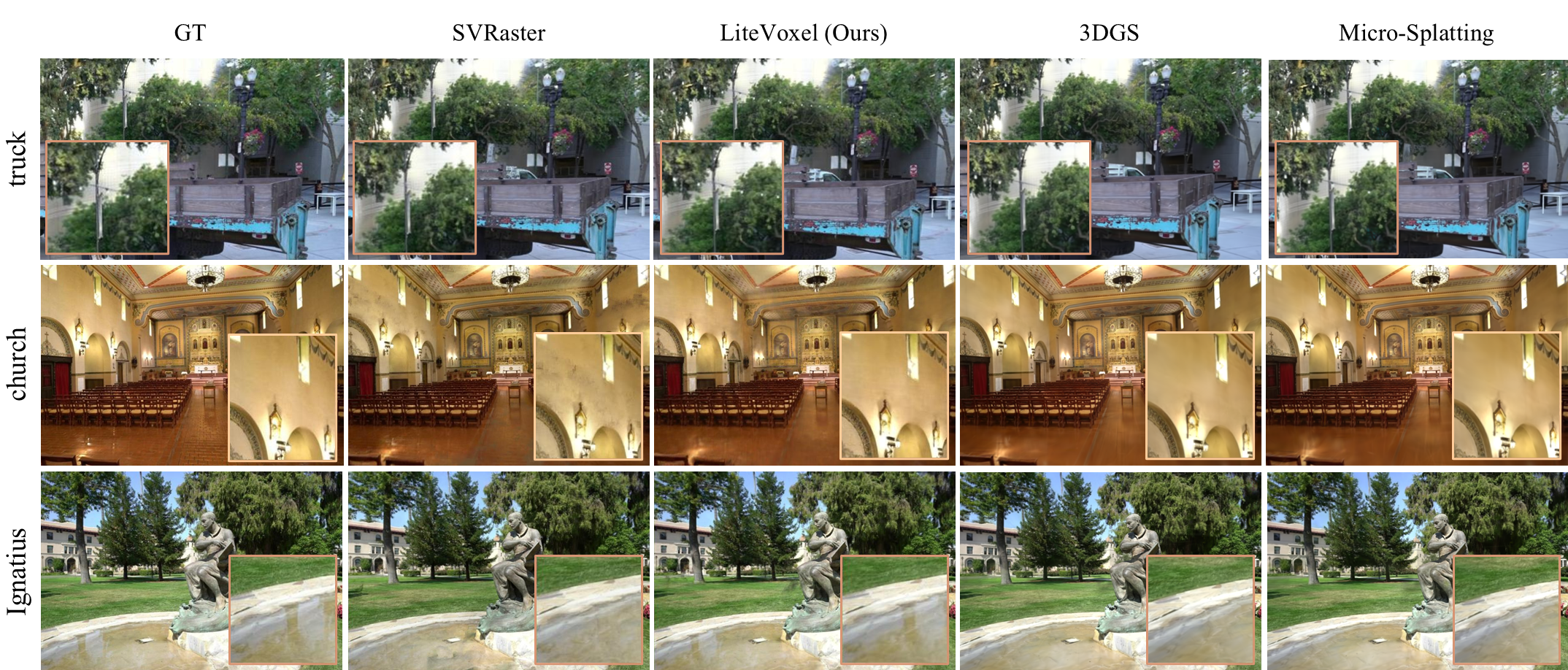}\\[2ex]
  {\footnotesize
  \parbox{\textwidth}{
    \textbf{Fig.~7. \textit{Qualitative comparison on Tanks \& Temples~\cite{ref20}.}} 
    (From left to right) Each row shows the ground truth, SVRaster~\cite{ref6}, LiteVoxel (Ours), 3DGS~\cite{ref2}, and Micro-Splatting~\cite{ref8}. 
    Scenes such as \textit{truck}, \textit{church}, and \textit{Ignatius} demonstrate how our approach captures detail 
    and removes errors more effectively.
  }}
\end{figure*}

\noindent
\\\\\\\\\indent SVRaster and LiteVoxel are trained with 20k iterations; 3DGS and Micro-Splatting are trained with 30k iteration budgets on a single RTX~5090~GPU. 
This identical compute/iteration budget ensures that any efficiency gains (e.g., memory, stability) arise from policy changes rather than extra optimization steps.
We report PSNR~\cite{ref21}, SSIM~\cite{ref11}, and LPIPS~\cite{ref22} on held-out views, along with efficiency metrics such as peak VRAM, final voxel/splat count, training time, and render FPS, to capture both quality and resource usage. 

\vspace{0.5em}
\noindent\textit{Quantitative Results.} 
Across Mip-NeRF~360, LiteVoxel achieves image quality on par with SVRaster and the Gaussian baseline while substantially lowering memory pressure, as shown in Fig.6 and Table 1. 
The peak VRAM is reduced by approximately $40\%\!-\!60\%$ on average without degrading PSNR, SSIM, or LPIPS, and FPS and training time remain similar to the baseline. 
These trends persist on Tanks~\&~Temples (Fig.7), where large-scale geometry and extended depth ranges amplify the benefits of depth-aware pruning and footprint-tested subdivision. 
The combined effect is a more compact model with fewer live voxels, steady silhouettes, and balanced refinement across depth, avoiding the myopic near-camera growth typical of global-threshold SVRaster. 
Detailed quantitative results for each scene are included in Appendix~I.

\vspace{0.6em}
\noindent\textit{Qualitative Analysis.} 
Figures on both datasets reveal characteristic failure modes and how LiteVoxel addresses them:

\begin{itemize}
  \item \textit{Low-frequency regions.} On Mip-NeRF 360 indoor scenes and outdoor flat regions, LiteVoxel suppresses blotchy residuals and evens out exposure and tonal drift while maintaining edge sharpness. This stems from the mid-training inverse-Sobel reweighting, which redirects gradient budget to flat regions only after geometry settles, avoiding over-smoothing of contours. The effectiveness of blotch suppression without softening edges is visible in Fig.~6 \textit{bonsai, counter,} and \textit{room} scenes and in Fig.~7 \textit{church} scene. 

  \item \textit{Silhouette stability.} Around thin structures and object boundaries, the EMA + hysteresis and keep-halo guards prevent flip-flop behavior and preserve delicate geometry during pruning passes, reducing flicker and “halo” artifacts. Fig.~6 \textit{bicycle} scene and Fig.~7 \textit{truck, Ignatius} shows how edges and silhouettes are preserved. 

  \item \textit{Depth fairness.} On Tanks \& Temples, far-depth details (distant facades, statues) are no longer neglected; a depth-aware quantile allocates prune pressure fairly across depth while footprint eligibility forbids over-refining regions the camera cannot resolve, producing cleaner far-field reconstructions without unnecessary voxel growth near the sensor. This can be observed in Fig.~7 \textit{truck, church} scenes where the details such as lightbulbs and ceiling edges are refined. 
\end{itemize}

\noindent These observations align with our stated goals to recover LF fidelity, stabilize sparsity updates, and bound growth without sacrificing high-frequency detail.

\vspace{1\baselineskip}

\noindent\textit{Ablation Study.} We ablate the three pillars of LiteVoxel (i) the LF-aware curriculum, (ii) the priority-driven subdivision with footprint eligibility, and (iii) the depth-stratified pruning with stability guards, by disabling one module at a time across Mip-NeRF 360 datasets (6scenes). Table II reports scene-level averages from the same run with evaluation metrics and full per-scene numbers and qualitative results are provided in the Appendix I. 

\vspace{1em}
\begin{figure*}[b]
  \centering
  \includegraphics[width=\textwidth]{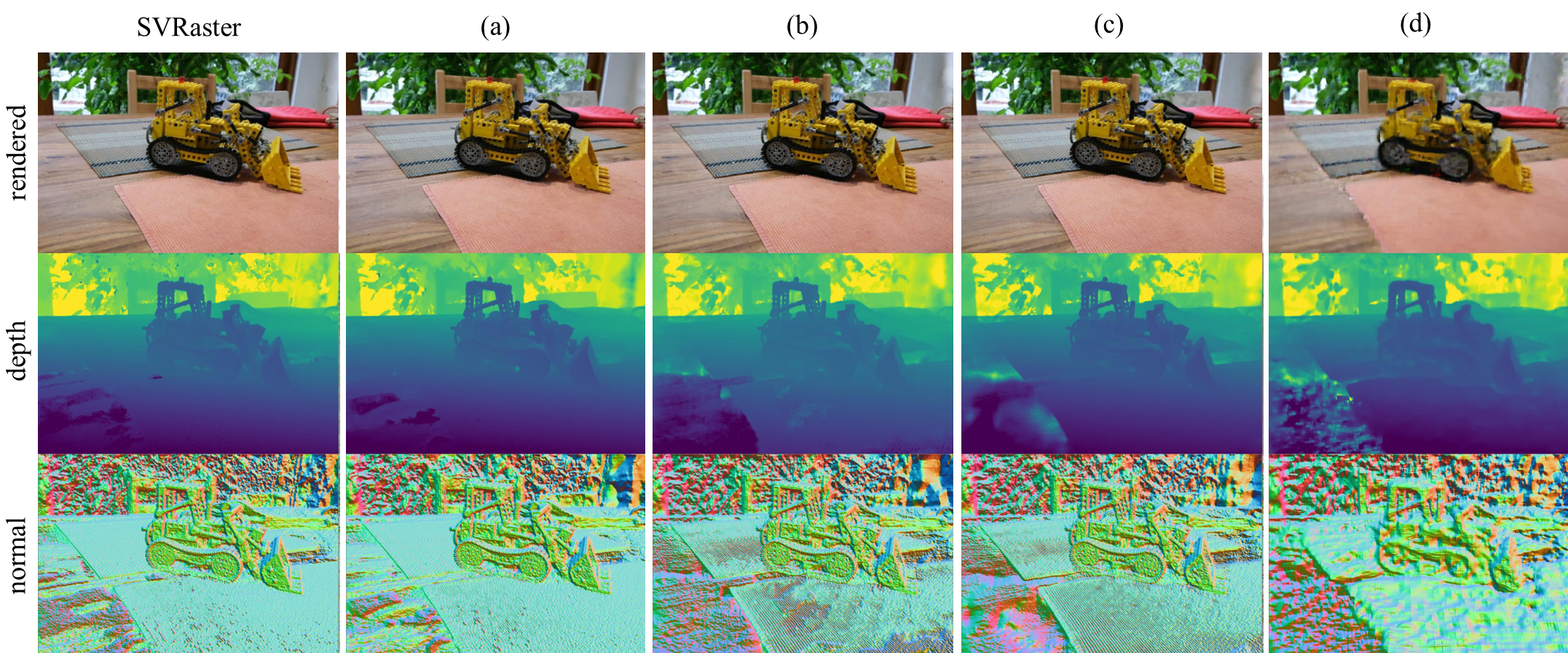}\\[2ex]
  {\footnotesize
  \parbox{\textwidth}{
    \textbf{Fig.~8.} Ablation on Mip-NeRF~360~\cite{ref19}: rendered, depth, and normal comparisons.
    Columns (left$\rightarrow$right): SVRaster (context baseline), 
    (a) Full LiteVoxel, 
    (b) -LF curriculum (LF reweighting off), 
    (c) -Pruning logic (no depth-quantile pruning or stability guards), 
    (d) -Priority-driven subdivision removed (no footprint/priority). 
    Rows show the final RGB render, depth, and surface normals under the same training budget and camera settings.
  }}
\end{figure*}

\vspace{-0.5em}  

\begin{itemize}
  \item (a) Full LiteVoxel. 
  \\\indent Fig.~8~(a) shows clean, uniform flats (tabletop/placemat),
  crisp edges on the bulldozer silhouette, and stable depth/normal fields without banding.
  Table~II confirms a balanced operating point with strong perceptual quality
  (SSIM~0.937, LPIPS~0.0652), competitive PSNR~(32.13), bounded peak VRAM~(7.9~GB)
  and compact model size~(4.67~M voxels) with real-time FPS~(307) and moderate training
  time~(7m~6s). This is the reference we aim to match in fidelity while enforcing memory predictability.

  \item (b) -- LF curriculum. 
  \\\indent Without the mid-training LF schedule, optimization remains edge-dominated.
  In Fig.~8~(b), flat regions (tabletop, placement) exhibit faint blotchy grain;
  depth shows mild low-frequency banding, and normal expose micro-speckle aligned
  with texture. Quantitatively, PSNR rises slightly~(32.58) but SSIM drops~(0.925)
  and LPIPS degrades~(0.0729)~(Table~II), which is a classic sign that the solution
  sharpens edges while sacrificing low-frequency smoothness. Resource usage remains
  similar~(8.0~GB, 4.64~M voxels, 7m~14s). The LF curriculum primarily improves
  flat-region fidelity at essentially no cost, which is not fully captured by PSNR alone
  but is visible in Fig.~8~(b) and detailed in Appendix~I.

  \item (c) -- Pruning logic. 
  \\\indent Disabling pruning removes the principal growth control. Fig.~8~(c) looks close at a glance, but the depth map thickens around object boundaries and normal show noisier micro-structure/halos which are symptoms of unbounded voxel acceleration. Table~II shows the cost: voxels jump to~8.04~M, peak memory to~12.3~GB, FPS drops to~238, and time increases to~9m~47s. The small PSNR bump~(32.77) is a diminishing-returns gain bought by excessive capacity; perceptual gains are negligible. Pruning with stability guards is the dominant efficiency lever as it keeps memory and model size predictable without hurting quality.

  \item (d) -- Priority-driven subdivision. \\\indent Removing footprint eligibility and depth-aware priority prevents the system from putting detail where the camera can resolve it. The result is under-refinement of essential regions and eroded thin structures (see Appendix~II zooms); normal exhibit regular cross-hatch artifacts. Table~II records the failure: quality collapses even though efficiency looks great on paper. Capacity allocation is not optional, as without the priority/footprint test, the model is “efficiently wrong”.
  \end{itemize}

  \begin{table}[!t]
  \centering
  \caption{Ablation Study Quantitative Analysis \\ (Averaged Across 6 Scenes)}
  \vspace{1ex}
  \includegraphics[width=0.85\linewidth]{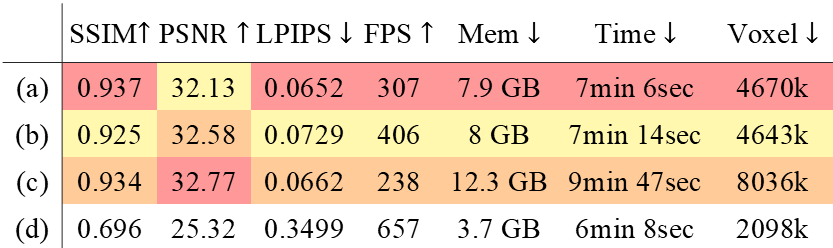}
\end{table}

\vspace{1\baselineskip}

For scene-by-scene quantitative tables across 7 scenes in Mip NeRF 360 dataset and additional qualitative crops that exhibit the above effects more clearly, please refer to the Appendix II.

\section{Discussion and Conclusion}
\noindent\textit{Discussion.} LiteVoxel reframes voxel rasterization as a memory-bounded, stability-constrained optimization regime, and the evidence across Mip-NeRF 360~\cite{ref19}.  and Tanks \& Temples~\cite{ref20} shows that this shift of not having heavier models or longer training, drives the outcomes. The ablations clarify the role of each pillar: the LF curriculum repairs low-frequency bias without taxing resources; depth-quantile pruning with EMA/hysteresis stabilizes sparsity kinetics and keeps VRAM predictable; and priority-driven subdivision allocates capacity only where the imaging footprint can resolve additional detail. Together they yield compact models with steady silhouettes and preserved high-frequency detail at matched fidelity. The approach does introduce schedule choices (e.g., LF activation window) and assumes non-pathological view coverage for reliable depth statistics; however, our guards (per-step caps, halo protection) mitigate most instability modes. Practically, LiteVoxel’s bounded-peak-VRAM behavior makes explicit rasterization viable on commodity GPUs for large scenes, enabling reproducible training budgets and easier deployment.

\vspace{1\baselineskip}

\noindent\textit{Conclusion.} We presented LiteVoxel, a principled training and adaptation framework for voxel rasterization that reallocates optimization effort and representational capacity under an explicit memory budget. On two challenging benchmarks, LiteVoxel matches or exceeds the visual quality of strong baselines while cutting peak VRAM and stabilizing topology evolution. Ablations confirm that each component is necessary and complementary: curriculum improves low-frequency fidelity, guarded pruning governs growth and stability, and priority-based subdivision focuses detail where it matters. Beyond the empirical gains, the key contribution is a predictable operating regime for explicit scene representations—one that scales to complex environments without trading away fidelity. Future work will explore auto-tuning of the schedules/quantiles, uncertainty-aware capacity rules, and extending the regime to dynamic scenes and multi-task supervision (e.g., normals/semantics) under the same memory constraints.

\clearpage
\onecolumn
\renewcommand{\thesection}{\Roman{section}}
\renewcommand{\thetable}{A\arabic{table}}
\setcounter{table}{0}

\section*{APPENDIX I.\ Per-Scene Quantitative Results}

\vspace{1\baselineskip}
This appendix reports the full per-scene metrics that underpin Sec.~V: \textit{Results and Evaluation (Quantitative Results)}. 
For each scene, we list PSNR, SSIM, LPIPS (held-out views) and efficiency statistics—peak VRAM, final voxel/splat count, render FPS, and training time—for the four methods compared in the main text: SVRaster, LiteVoxel (ours), 3D Gaussian Splatting (3DGS), and Micro-Splatting. 
All numbers were obtained under an identical iteration budget, camera intrinsics/poses, and render settings. 
We use a color rank legend (red = best, orange = second, yellow = third) to make scene-wise winners immediately visible. 
Scene-level tables are provided for the full set of Mip-NeRF 360 scenes and the Tanks \& Temples subset used in Sec.~V; dataset means that appear in the main text can be recomputed directly from these rows. 
Cropped qualitative comparisons already appear in Sec.~V; readers may cross-reference them with these per-scene tables for a complete view of quality–efficiency trade-offs.

\vspace{2em}

\begin{center}
  {\normalfont TABLE AI\\ QUANTITATIVE ANALYSIS ON MIP-NERF 360 DATASET}\\[0.5em]
  \includegraphics[width=\textwidth,height=0.45\textheight,keepaspectratio]{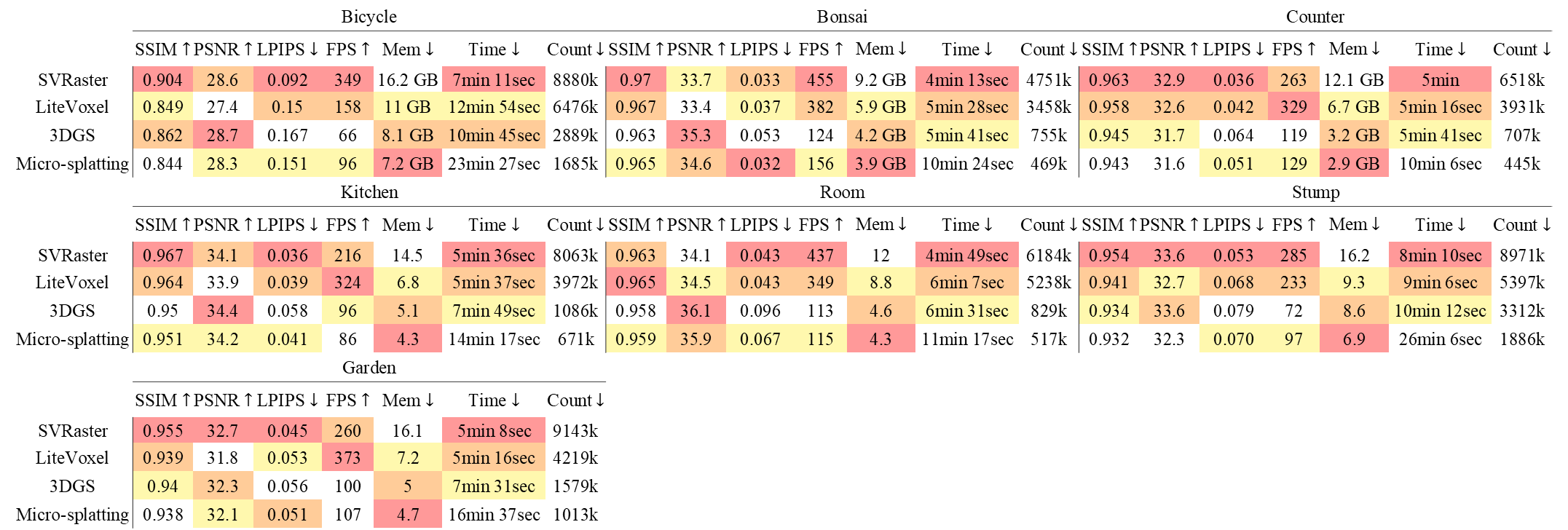}\\[0.5em]
\end{center}

\vspace{3em}

\begin{center}
  {\normalfont TABLE AII\\ QUANTITATIVE ANALYSIS ON TANKS \& TEMPLES DATASET}\\[0.5em]
  \includegraphics[width=\textwidth,height=0.45\textheight,keepaspectratio]{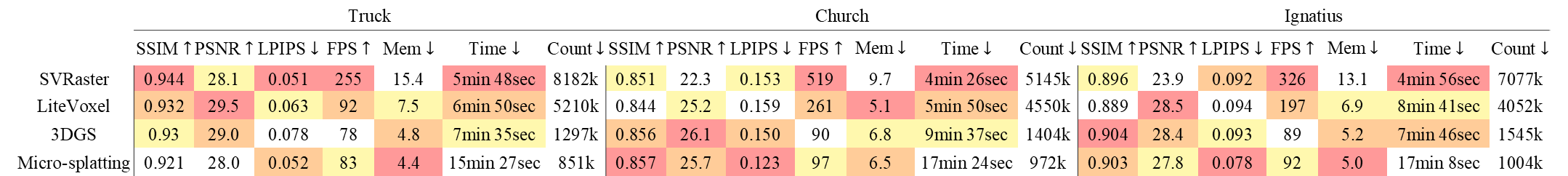}\\[0.5em]
\end{center}

\vspace*{-1em}
\clearpage
\onecolumn
\renewcommand{\thefigure}{A\arabic{figure}} 
\setcounter{figure}{0}

\section*{APPENDIX II.\ Ablation Study Details}

\vspace{1\baselineskip}
\noindent
This appendix complements Sec.~V\textsubscript{D} by reporting per-scene results for the ablations on Mip-NeRF~360 (Tanks \& Temples ablations were not performed). 
We keep the same labels as the main text and restate them here for clarity: 
(a) Full LiteVoxel (ours), 
(b) LF curriculum (inverse-Sobel reweighting disabled), 
(c) Pruning logic (depth-quantile pruning and stability guards disabled), 
(d) Priority-driven subdivision (footprint eligibility / priority allocator disabled). 
Figure~A1 presents cropped GT vs.~(a)–(d) for each scene; 
Tables~A3 (PSNR/SSIM/LPIPS) and~A4 (FPS, peak VRAM, training time, voxel count) list the corresponding numbers.

\vspace{2em} 

\begin{center}
  \includegraphics[width=\textwidth]{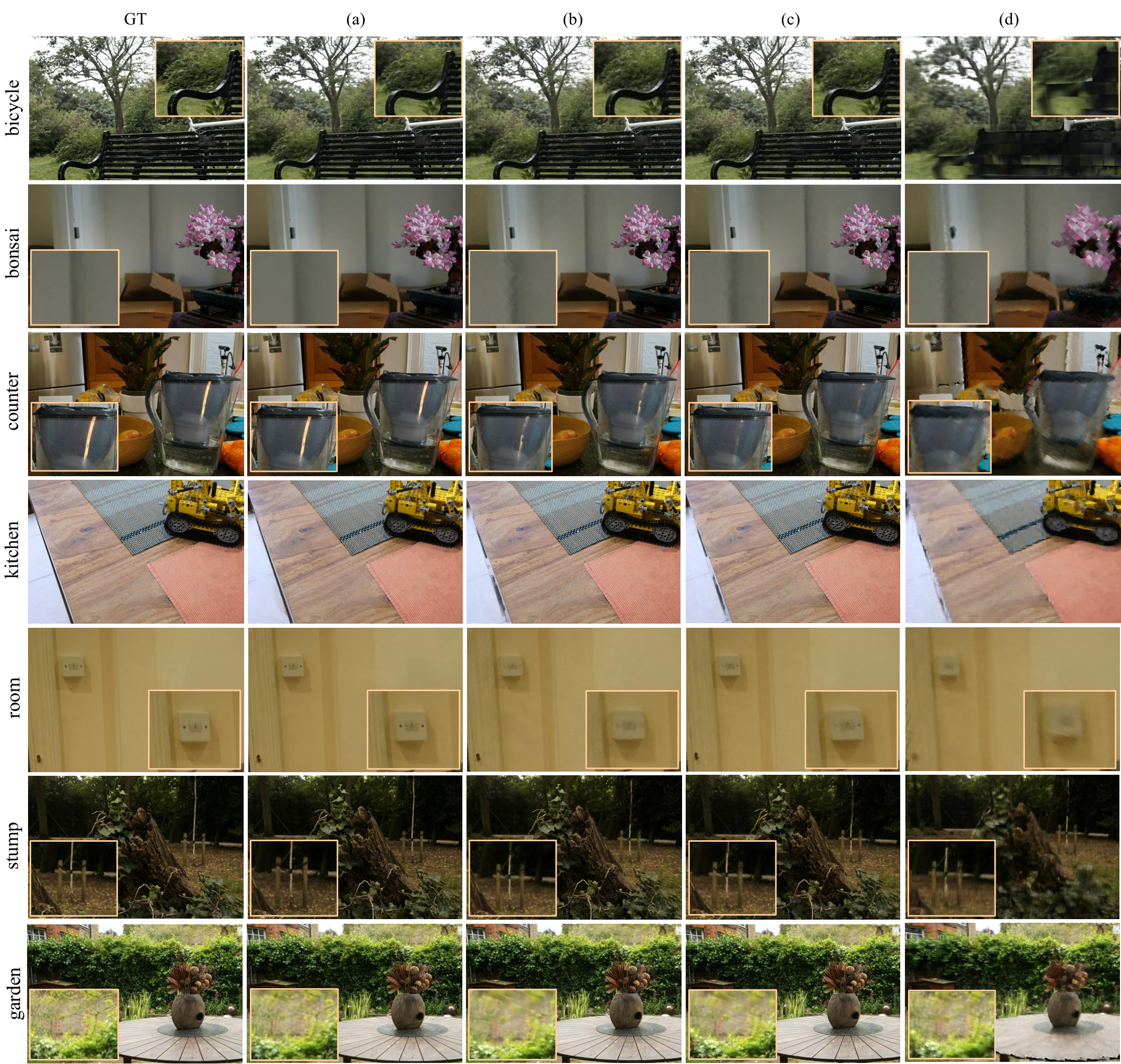}\\[2ex]
  {\footnotesize
  \parbox{\textwidth}{
    \textbf{Fig.~A1.} Ablation study on Mip-NeRF~360—per-scene qualitative comparisons. 
    Columns (left → right): GT, (a) Full LiteVoxel, (b) -LF curriculum (inverse-Sobel reweighting disabled), 
    (c) Pruning logic (no depth-quantile pruning or stability guards), 
    (d) Priority-driven subdivision removed (no footprint/priority). 
    Rows are 7 scenes from the Mip-NeRF~360 dataset—bicycle, bonsai, counter, kitchen, room, stump, and garden.
  }}
\end{center}

\indent In Fig. A1 the qualitative trends are consistent across scenes. The --LF curriculum variant (b) leaves flat regions with fine blotchy grain and faint banding--clearly visible on the walls and planar patches in room, counter, kitchen, and the cardboard in bonsai--whereas the full LiteVoxel model (a) suppresses these artifacts while keeping edges crisp. The --priority subdivision variant (d) misallocates capacity: foliage in bicycle and background structure in garden appear blocky or eroded, and thin details go missing, a hallmark of ignoring footprint/priority during splitting. By contrast, removing pruning (--pruning, c) permits uncontrolled growth; textures may look marginally sharper at a glance, but boundary halos and micro-noise increase, as seen on stump bark and the bicycle rail.

\begin{center}
  {\normalfont TABLE AIII\\ MIP-NERF 360 ABLATION (PER-SCENE): IMAGE-QUALITY METRICS}\\[0.5em]
  \includegraphics[width=\textwidth,height=0.45\textheight,keepaspectratio]{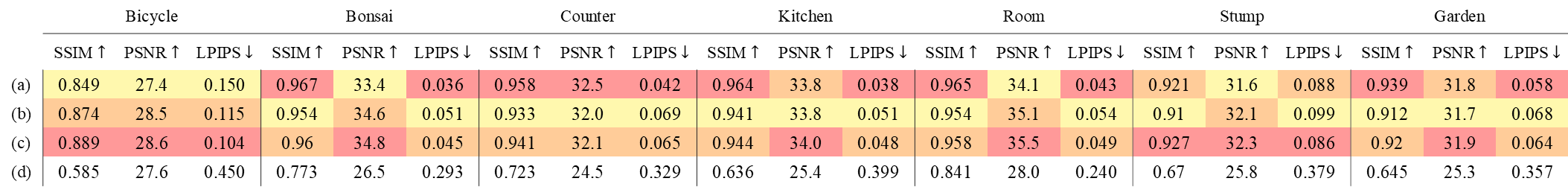}\\[0.5em]
\end{center}

\begin{center}
  {\normalfont TABLE AIV\\ MIP-NERF 360 ABLATION (PER-SCENE): EFFICIENCY METRICS}\\[0.5em]
  \includegraphics[width=\textwidth,height=0.45\textheight,keepaspectratio]{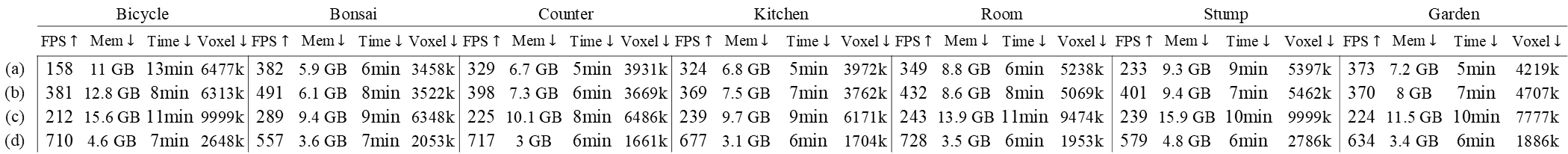}\\[0.5em]
\end{center}

Tables A3 and A4 corroborate these observations quantitatively. In Table A3, (b) often shows a small PSNR uptick yet worse LPIPS/SSIM (e.g., room, counter), matching the “sharper but blotchier” appearance; (c) sometimes attains the highest PSNR but negligible perceptual gains, while (d) collapses across all three quality metrics. Table A4 shows that (c) inflates voxel count and peak VRAM and reduces FPS with longer training time—evidence that pruning with guards is the main efficiency lever. Although (d) looks fast and light on paper (high FPS, low VRAM), its severe quality drop in A3 confirms it is efficiently wrong. Full LiteVoxel (a) strikes the intended balance: low LPIPS and strong SSIM at bounded VRAM with compact voxel counts.
\\\indent These per-scene ablations substantiate the claims in Sec. V-D: the LF curriculum fixes flat-region bias at negligible cost; guarded pruning stabilizes sparsity and memory; and priority-driven subdivision is essential to place detail where the camera can resolve it.

\end{document}